\documentclass[conference]{IEEEtran}
\IEEEoverridecommandlockouts
\usepackage{cite}
\usepackage{amsmath,amssymb,amsfonts}
\usepackage{algorithmic}
\usepackage{graphicx}
\usepackage{textcomp}
\usepackage{xcolor}
\usepackage{booktabs}
\usepackage{textgreek}
\def\BibTeX{{\rm B\kern-.05em{\sc i\kern-.025em b}\kern-.08em
    T\kern-.1667em\lower.7ex\hbox{E}\kern-.125emX}}
\begin{document}

\title{M3ET: Efficient Vision-Language Learning for Robotics based on Multimodal Mamba-Enhanced Transformer\\

\thanks{\textsuperscript{*} Corresponding author.}
}

\author{
\IEEEauthorblockN{1\textsuperscript{st} Yanxin Zhang}
\IEEEauthorblockA{\textit{School of Software} \\
\textit{Northwestern Polytechnical University}\\
Xi'an, China \\
zhangyanxin2024@mail.nwpu.edu.cn}
\and
\IEEEauthorblockN{2\textsuperscript{nd} Liang He}
\IEEEauthorblockA{\textit{School of Software} \\
\textit{Northwestern Polytechnical University}\\
Xi'an, China \\
2021050018@nwpu.edu.cn}
\and
\IEEEauthorblockN{3\textsuperscript{rd} Zeyi Kang}
\IEEEauthorblockA{\textit{School of Software} \\
\textit{Northwestern Polytechnical University}\\
Xi'an, China \\
kangzeyi@mail.nwpu.edu.cn}
\and
\IEEEauthorblockN{4\textsuperscript{th} Zuheng Ming\textsuperscript{*}}
\IEEEauthorblockA{\textit{Laboratoire L2Tl} \\
\textit{University Sorbonne Paris Nord}\\
Paris, France \\
zuheng.ming@univ-paris13.fr}
\and
\IEEEauthorblockN{5\textsuperscript{th} Kaixing Zhao\textsuperscript{*}}
\IEEEauthorblockA{\textit{School of Software} \\
\textit{Yangtze River Delta Research Institute (Taicang)} \\
\textit{Northwestern Polytechnical University}\\
Xi'an, China \\
kaixing.zhao@nwpu.edu.cn}
}

\maketitle

\begin{abstract}
In recent years, multimodal learning has become essential in robotic vision and information fusion, especially for understanding human behavior in complex environments. However, current methods struggle to fully leverage the textual modality, relying on supervised pretrained models, which limits semantic extraction in unsupervised robotic environments—particularly with significant modality loss. These methods also tend to be computationally intensive, leading to high resource consumption in real-world applications. To address these challenges, we propose the Multi-Modal Mamba-Enhanced Transformer (M3ET), a lightweight model designed for efficient multimodal learning, particularly on mobile platforms. By incorporating the Mamba module and a semantic-based adaptive attention mechanism, M3ET optimizes feature fusion, alignment, and modality reconstruction. Our experiments show that M3ET improves cross-task performance, with a 2.3 times increase in pre-training inference speed. Particularly, the core VQA task accuracy of M3ET remains at 74.18\%, while the model’s parameter count is reduced by 66.63\%. Although performance on the EQA task is limited, M3ET's lightweight design makes it well-suited for deployment on resource-constrained robotic platforms.
\end{abstract}


\section{Introduction}
Human-Robot Interaction (HRI) relies on accurate environmental understanding, which has garnered significant attention \cite{ref2}. Robots, equipped with perception capabilities, can perform tasks like obstacle avoidance \cite{ref38} and path planning \cite{ref39}, essential for HRI. Multimodal learning has proven valuable in HRI by integrating complementary information from different modalities, enhancing scene representations and decision-making \cite{ref40}. However, balancing rich semantic information with computational constraints in real-world robot deployment remains challenging \cite{ref41}.

Masked Autoencoders (MAEs) \cite{ref3} are effective self-supervised methods, particularly for Vision Transformers (ViTs) \cite{ref1}, masking image patches and reconstructing missing regions. While MAEs are effective for image data, they fail to fully utilize other modalities like depth or semantic data \cite{ref42}, which is crucial in real-world HRI scenarios with multimodal sensors. Early multimodal methods often focus on pairwise relationships, but they overlook deeper cross-modal semantics that could improve performance \cite{ref43}. Transformer-based models \cite{ref6}, while good at modeling long-range dependencies, suffer from large sizes and high computational costs, hindering deployment in resource-constrained environments.

To address these issues, we propose M3ET, a lightweight multimodal framework integrating semantic information through a specialized fusion module. Unlike traditional methods that stack textual embeddings onto image or depth features, M3ET aligns semantic signals with sensory data, enhancing feature discrimination while maintaining efficiency for edge devices. Our method exploits semantic information \cite{ref12} for improved reconstruction and recognition, offering a balance between performance and compactness for real-time applications.

We evaluate M3ET on standard multimodal datasets \cite{ref13}, comparing it with state-of-the-art methods. Our results show significant improvements in reconstruction quality, especially in challenging scenarios. Ablation studies confirm that incorporating semantic modalities does not compromise performance. Our contributions are:

1. A unified multimodal framework integrating semantic priors from large language models into a lightweight architecture, achieving competitive accuracy even under high masking conditions.

2. Enhanced interpretability through better interaction between high-level semantic and low-level sensory features, improving decision-making while minimizing computational costs.

3. We conduct extensive evaluations across reconstruction, VQA, and EQA tasks, demonstrating M3ET’s generalization ability and robustness to modality degradation.

4. We show that M3ET's compact architecture and 2.3× speedup make it suitable for real-world robotic deployment on resource-limited edge devices.

The paper is organized as follows: Section II reviews related work, Section III details the proposed method, Section IV is experimental results, and Section V concludes with future directions.

\section{Related works}
We review advancements in multimodal learning, focusing on methods that address modality loss and the integration of semantic information from language models. We analyze existing models under extreme modality loss and explore how incorporating textual modalities enhances reconstruction. Additionally, we discuss multimodal learning in robotics \cite{ref13}, particularly in tasks like control \cite{ref44}, path planning \cite{ref15}, perception \cite{ref45}, and decision-making \cite{ref46}, where visual and textual data provide richer context. Finally, we highlight shortcomings of existing methods and explain how M3ET addresses these challenges.

\subsection{Multimodal Learning}
Existing multimodal methods mainly focus on visual and textual data relationships. MultiMAE \cite{ref5} is a self-supervised method that masks parts of an image and uses unmasked portions for reconstruction. While MultiMAE excels at learning visual features, it mainly relies on image modalities and does not fully utilize textual data, a limitation in real-world settings where data is often incomplete \cite{ref16}.

Models like ViLBERT \cite{ref17} and CLIP \cite{ref18} combine vision and language but suffer from high computational costs, limiting their use in real-time robotic applications. These models are effective for specific tasks but struggle in resource-constrained environments, where modality loss and real-time processing are frequent.

M3ET addresses these challenges by integrating semantic information from language models, enhancing multimodal learning, especially when modality loss occurs. M3ET optimizes visual and textual feature fusion while maintaining computational efficiency, ensuring robust performance in real-time robotic applications like human-robot interaction.

\subsection{Lightweight Models for Robotics}
In robotics, lightweight models are essential, especially in embodied applications with limited computational resources. While models like TinyGPT-V \cite{ref19} and LLaVA-Phi \cite{ref20} reduce model size and computational load, they fail to address challenges like modality loss and efficient multimodal processing in dynamic environments.

Efficient lightweight architectures are critical for real-time performance in robotics. However, prior models typically focus on individual modalities and fail to integrate visual and textual data to compensate for missing information. These models are often designed for general tasks, not optimized for robotic decision-making.

To address this, we propose M3ET, a lightweight architecture combining Mamba's efficiency with deep semantic information from text. Unlike existing models, M3ET integrates textual data to complement missing visual data, enhancing multimodal task performance while maintaining a lightweight design.

\section{Methodology}

In this section, we describe the structure of the Multi-Modal Mamba-Enhanced Transformer (M3ET) and its training procedures. First, we introduce the M3ET model architecture, focusing on how the Mamba module integrates with the Transformer architecture to enhance computational efficiency and long-range dependency modeling. We also present the multi-modal sampling and masking strategy in Section \ref{sec:training-strategy}, which ensures balanced token distribution across modalities to optimize masked autoencoding \cite{ref23} and improve training stability and generalization.

\subsection{Model Architecture}\label{sec:molel-architecture}

\begin{figure*}[htbp]
    \centering
    \includegraphics[width=\linewidth]{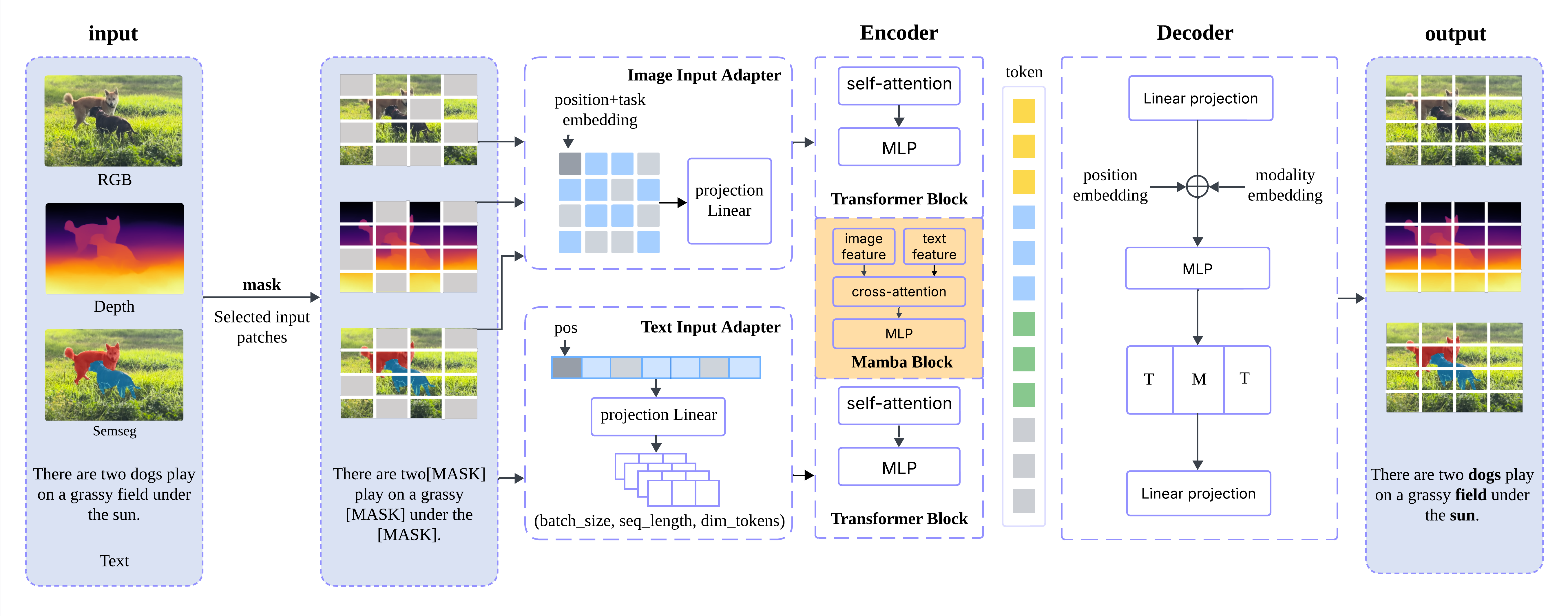}
    \caption{\small \textbf{Multi-Modal Mamba-Enhanced Transformer architecture:}The framework primarily consists of multimodal data at the input end, image
adapter, text adapter, encoder, decoder, and output end. The input adaptation module converts RGB, depth, and semantic segmentation (Semseg) images
through the vision adapter, while the text is tokenized via the language adapter to generate unified modal embeddings. The cross-modal encoder employs
alternating Transformer and Mamba blocks to hierarchically fuse visual and textual features, with the Mamba module efficiently modeling long-range
dependencies. The decoder reconstructs images and generates text through modality-specific heads, using shared cross-attention layers to align multimodal
representations. The model performs multimodal tasks by leveraging multimodal information, ensuring efficient feature interaction and task specialization.}
    \label{fig:d_image}
\end{figure*}

Since traditional Transformer-based models often suffer from large parameter sizes and high computational complexity (which make them inefficient for deployment on resource-constrained platforms such as robots), Mamba offers a lightweight alternative by reducing the number of parameters and computational overhead without compromising the model's ability to capture rich semantic relationships \cite{ref27}. Based on this idea, our proposed M3ET framework employs a hierarchical architecture, as shown in Figure \ref{fig:d_image}. More precisely, the Mamba module integrates State Space Modeling (SSM) and Selective SSM to improve efficiency, particularly in sequence processing tasks.

\textbf{State Space Models (SSM):}
SSMs describe dynamic systems. For a one-dimensional input signal \( x(t) \in \mathbb{R} \), the model maps \( x(t) \) to output \( y(t) \) through a latent state \( h(t) \):

\[
h'(t) = A h(t) + B x(t) \tag{1}
\]
\[
y(t) = C h(t) \tag{2}
\]

The SSM is discretized into a convolution operation for parallel computation:

\[
K = (C B, C A B, \dots, C A^{L-1} B), \quad y = x \cdot K \tag{3}
\]

\textbf{Selective SSM:}  
Selective SSM introduces input-dependent parameters:

\[
B = s_B(x), \quad C = s_C(x), \quad \Delta = \tau_\Delta(P + s_\Delta(x)) \tag{4}
\]

This allows dynamic parameter selection and improves long-range dependency modeling.

\textbf{State Space Duality:}  
Mamba connects structured SSM with Transformer attention mechanisms. The dual structure reformulates the SSM transformation for better efficiency:

\[
y = \text{SSM}(A, B, C)(x) \tag{5}
\]
\[
M_{ji} = C^T \prod_{k=i}^j A_k B_i \tag{6}
\]
\[
M = L \cdot (C B^T) \tag{7}
\]

where \( x \) represents the input feature from any modality. This dual structure allows efficient multimodal handling. \( L \) is a lower-triangular mask matrix, enabling attention-like computation.

The proposed framework significantly reduces computational overhead while maintaining the model's capacity for long-range dependency modeling and context-based reasoning, which is critical for efficient multimodal learning in robotics.

The MambaBlock transformation is as follows:
\[
X' = \text{LayerNorm}(X)
\tag{8}
\]
which applies layer normalization to the input features \( X \). Here, layer normalization ensures that the input features are standardized before being passed through the network.

In addition, feed-forward step is applied using a two-stage transformation:

\[
H = W_2 \cdot \text{GELU}(W_1 X')
\tag{9}
\]
Here, \( W_1 \in \mathbb{R}^{768 \times 64} \) and \( W_2 \in \mathbb{R}^{64 \times 768} \) are learnable weight matrices. The input \( X' \) refers to the feature vector obtained after the Layer Normalization step, and it could originate from images or text. During the processing, the input \( X' \) is first transformed by \( W_1 \) into a lower-dimensional space (64 dimensions), followed by applying the GELU activation function, which introduces non-linearity. Then, \( W_2 \) projects the result back to the original dimensionality (768 dimensions).

Finally, the output is regularized using dropout:

\[
H_{\text{output}} = \text{Dropout}(H)
\tag{10}
\]
Which applies on transformed features \( H \) to prevent overfitting during training by randomly deactivating some of the activations.

In general, the M3ET framework incorporates three specialized decoders for different tasks: depth estimation, text generation, and RGB reconstruction. Each decoder employs a unique computational structure optimized for its task. Every decoder begins with three Transformer layers, processing local features through self-attention with 256-dimensional QKV projections. A critical MambaBlock is integrated into the fourth layer, compressing spatial and textual representations from 256 to 64 dimensions and restoring them to 256 dimensions. This step enhances the modeling of long-range dependencies across modalities. Two additional Transformer layers refine the extracted features to improve task-specific predictions.

The cross-attention mechanism is a key component in multi-modal models, enabling the interaction between different modalities\cite{ref29}. It allows the model to selectively attend to relevant features from one modality while considering the context provided by another modality. M3ET model incorporates a dynamic fusion mechanism to facilitate interactions between different modalities. Modality-specific embeddings are stored within a parameter dictionary of 256 dimensions, preserving the unique characteristics of each modality. These embeddings interact through learnable projection matrices, governed by the cross-attention mechanism:
\begin{equation}\small
\operatorname{Atten}(Q_r, K_d, V_t) = \operatorname{softmax}\left( \frac{Q_r W_q (K_d W_k)^{\top}}{\sqrt{d}} \right) V_t W_v
\tag{11}
\end{equation}

Where:
\begin{itemize}
    \small
    \item \( Q_r \): Query matrix for the RGB image 
    \item \( K_d \): Key matrix for the depth image
    \item \( V_t \): Value matrix for the text modality
    \item \( W_q, W_k, W_v \in \mathbb{R}^{768 \times 256} \): Projection matrices that align visual and linguistic features
    \item \( d \): Dimension of the key vectors, used for scaling the dot product in attention
\end{itemize}

Mamba as a structured State Space Model (SSM), offers efficient long-range sequence modeling with linear complexity, making it well-suited for multimodal tasks involving image patches and text tokens. Unlike Transformers that use global attention, Mamba introduces selective gating to dynamically update modality-specific features. This improves fusion under modality loss while reducing redundant computation. Integrated hierarchically within M3ET, Mamba enables dynamic computation allocation, allowing the model to maintain strong performance across modalities with lower overhead—ideal for real-time robotic applications like VQA.

\subsection{Loss Function and Training Strategy}\label{sec:training-strategy}

\textbf{Loss Function Design:}
The total loss combines task-specific reconstruction losses and semantic alignment loss as follows:

\[
L_{\text{total}} = \lambda_{\text{rgb}} L_{\text{rgb}} + \lambda_{\text{dep}} L_{\text{dep}} + \lambda_{\text{seg}} L_{\text{seg}} + \lambda_{\text{text}} L_{\text{text}} 
\tag{12}
\]

In this equation, the total loss function $L_{\text{total}}$ is a weighted sum of different loss components for each modality and task. The weights $\lambda_{\text{rgb}}, \lambda_{\text{dep}}, \lambda_{\text{seg}},$ and $\lambda_{\text{text}}$ control the importance of each loss component.

\[
L_{\text{rgb}} = \frac{1}{N} \sum_{i=1}^{N} \mathbf{1}_{\text{mask}}(i) \cdot \left( \hat{y}(i) - y(i) \right)^2
\tag{13}
\]
$L_{\text{rgb}}$ is the loss for RGB image reconstruction, computed using Masked MSE to measure the error between the original and predicted RGB images. The weight $\lambda_{\text{rgb}}$ determines its contribution to the total loss.

\[
L_{\text{dep}} = \frac{1}{N} \sum_{i=1}^{N} \mathbf{1}_{\text{mask}}(i) \cdot \left| \hat{y}(i) - y(i) \right|
\tag{14}
\]
$L_{\text{dep}}$ is the loss for depth estimation, computed using Masked L1 to calculate the absolute error between predicted and true depth maps, with $\lambda_{\text{dep}}$ controlling its importance.

\[
L_{\text{seg}} = -\frac{1}{N} \sum_{i=1}^{N} \mathbf{1}_{\text{mask}}(i) \cdot \sum_{c=1}^{C} y_c(i) \log(\hat{y}_c(i))
\tag{15}
\]
$L_{\text{seg}}$ is the loss for semantic segmentation, computed using Masked Cross-Entropy to measure the difference between predicted and ground truth class labels, with $\lambda_{\text{seg}}$ indicating its contribution.

\[
L_{\text{text}} = - \sum_{i=1}^{N} y(i) \log(\hat{y}(i))
\tag{16}
\]
$L_{\text{text}}$ is the loss for text generation, computed using Cross-Entropy, ensuring the model generates accurate text, with $\lambda_{\text{text}}$ determining its weight in the total loss.

\textbf{Multi-modal Sampling and Masking Strategy:}
We propose a multi-modal sampling strategy to ensure balanced masking across modalities. First, we use the Dirichlet Distribution to sample the masking ratio, ensuring even distribution of masks. The number of sampled tokens is set to 98 (70\% of the total tokens).

For the visual modality, we use a spatially uniform masking strategy, randomly masking image patches with equal probability. For the text modality, we apply whole-sentence masking, with each sentence having an 80\% probability of being fully masked to enhance semantic consistency in text generation tasks\cite{ref10,ref30}. Additionally, we use a uniform sampling strategy during task selection to ensure equal probability across all modalities.

\section{Experiments}

We evaluate the proposed M3ET framework through extensive experiments, covering multimodal reconstruction, efficiency improvements, and real-world deployment. This section describes the datasets, baseline models, implementation details, and experimental setup, followed by evaluations on image-text reconstruction, VQA, and EQA tasks. A comprehensive ablation study analyzes the contributions of each component of the M3ET framework.

\subsection{Experimental Setup}

\textbf{Datasets:} We conducted experiments on four widely used datasets: ImageNet-1K, NYUv2, VQAv2, and OpenEQA. ImageNet-1K provides diverse images, and NYUv2 includes depth information for evaluating multimodal data with depth estimation. VQAv2 is a standard dataset for visual question answering tasks\cite{ref31}, and OpenEQA challenges models with open-domain, real-world questions. It consists of over 1,600 questions on topics like object recognition and spatial reasoning. We preprocessed ImageNet-1K and NYUv2 by generating descriptive texts from RGB images to enhance the text modality input.

To ensure efficient multimodal learning, we preprocess data differently for each modality. RGB and depth images are standardized using dataset-specific mean and variance. Depth images undergo truncated normalization to reduce noise. For text, we use the BERT tokenizer, followed by padding/truncation to 128 tokens, with fixed sine-cosine positional embeddings added. Visual modalities are split into non-overlapping 16×16 patches\cite{ref30}, while text tokens are projected into a 768-dimensional embedding space.

\textbf{Baselines:} We compare M3ET against state-of-the-art multimodal architectures, including BLIP, FLAVA, and MultiMAE with textual enhancements. For the VQA task, we benchmark M3ET against models like ViLBERT\cite{ref32}, LXMERT\cite{ref33}, and UNITER\cite{ref34}. For the OpenEQA dataset, we compare with existing methods in open-domain question answering tasks.

\textbf{Metrics:} We use PSNR(dB), SSIM, and LPIPS for image reconstruction tasks. Depth estimation is evaluated using RMSE, and semantic segmentation is measured using mean Intersection-over-Union (mIoU). For VQA, we report overall accuracy, and for OpenEQA, we measure performance using accuracy and F1-score. To evaluate efficiency, we report parameter count, FLOPs reduction, and inference speed.

\textbf{Implementation Details:} We pretrain M3ET using the ImageNet-1K dataset with the AdamW optimizer, a learning rate of 1e-4, and a batch size of 16 for 1600 iterations. The architecture consists of a 12-layer encoder with a hidden dimension of 768 and a 3-layer decoder. For fine-tuning on the VQA task, we use the VQAv2 dataset with a batch size of 32, an initial learning rate of 5e-5, and cosine annealing for 10 epochs. We use a similar setup for fine-tuning on the OpenEQA task.

\subsection{Evaluation on Multimodal Reconstruction and VQA}
We evaluate our M3ET framework on a variety of real and synthetic datasets, of varying size and complexity. We also conduct a comprehensive ablation study supporting our design choices.

\textbf{Multimodal Reconstruction Capability:} 
We compared M3ET with other multimodal models under the same training data and settings, focusing on image reconstruction performance. We trained with 2 Nvidia RTX 4090 GPUs. As shown in Table \ref{fig:2}, we report top-1 accuracy on ImageNet-1K (IN-1K) and $\sigma_1$ accuracy on the NYUv2 depth task. M3ET shows improvements over MultiMAE. On ImageNet-1K, M3ET's accuracy is 83.6\% (vs. 83.3\% for MultiMAE) and on NYUv2, M3ET performs better (86.8\%) than MultiMAE (86.3\%). The data shows that M3ET outperforms other methods on both datasets. The main limitation of the original framework is its linear token projection, which flattens both image and text modalities into homogeneous tokens before fusion, restricting semantic interactions. This design limits hierarchical alignment between textual concepts and image regions. This led us to adopt the Mamba multi-stage fusion architecture with separate modality encoders. Mamba's dynamic gating mechanism enhances image features at multiple scales, improving image reconstruction.

\begin{table}[ht]
\centering
\caption{\small Comparison of M3ET with other multimodal frameworks. We report the top-1 accuracy (↑) on ImageNet-1K (IN-1K), NYUv2 semantic segmentation(semseg) and δ1 accuracy (↑) on NYUv2 depth.}
\resizebox{\linewidth}{!}{
\begin{tabular}{lcccc}
\toprule
\textbf{Metric}          & \textbf{Supervised\cite{ref35}}  & \textbf{DINO\cite{ref36}}  & \textbf{MultiMAE}  & \textbf{M3ET} \\
\midrule
\textbf{ImageNet-1K (IN-1K)}     & 81.8              & 83.1               & 83.3               & \textbf{83.6}        \\

\textbf{NYUv2 (semseg)}& 50.1            & 47.9               & \textbf{52.0}               & \textbf{52.0}        \\

\textbf{NYUv2 (depth)} & 80.7            & 81.3               & 86.3               & \textbf{86.8}        \\

\bottomrule
\end{tabular}
}
\label{fig:2}
\end{table}

\textbf{VQA Performance:} For the Visual Question Answering (VQA) task, we fine-tuned M3ET on the VQAv2 dataset. As shown in Table \ref{tab:tab2}, M3ET achieved an accuracy of 74.18\% on the VQA validation set, surpassing previous models. Macro-Averaging calculates the average of metrics for each category, focusing on smaller categories, while Micro-Averaging performs a weighted average based on the number of samples, reflecting overall performance. M3ET achieved an overall accuracy of 73.62\%, indicating strong classification performance.

\begin{table}
\centering
\caption{\small Model performance Comparison on the down-stream VQA task}
\resizebox{\linewidth}{!}{
\begin{tabular}{l c c c c c}
\toprule
\textbf{Model} & \textbf{ViLBERT} & \textbf{LXMERT} & \textbf{UNITER} & \textbf{M3ET}\\
\midrule
Accuracy       & 70.55\%& 72.50\%& 68.19\%& \textbf{74.18\%}
\\
Macro-Averaging & 68.34\%& 69.18\%&\textbf{69.55\%}& 68.46\% \\
Micro-Averaging & 70.31\%& 72.93\%& 73.39\%& \textbf{73.62\%} \\
\bottomrule
\end{tabular}
}
\label{tab:tab2}
\end{table}
M3ET's performance demonstrates the effectiveness of multimodal pretraining in enhancing the model's ability to integrate visual and textual information, contributing to more accurate question answering.

\begin{figure}
    \centering
    \includegraphics[width=\columnwidth]{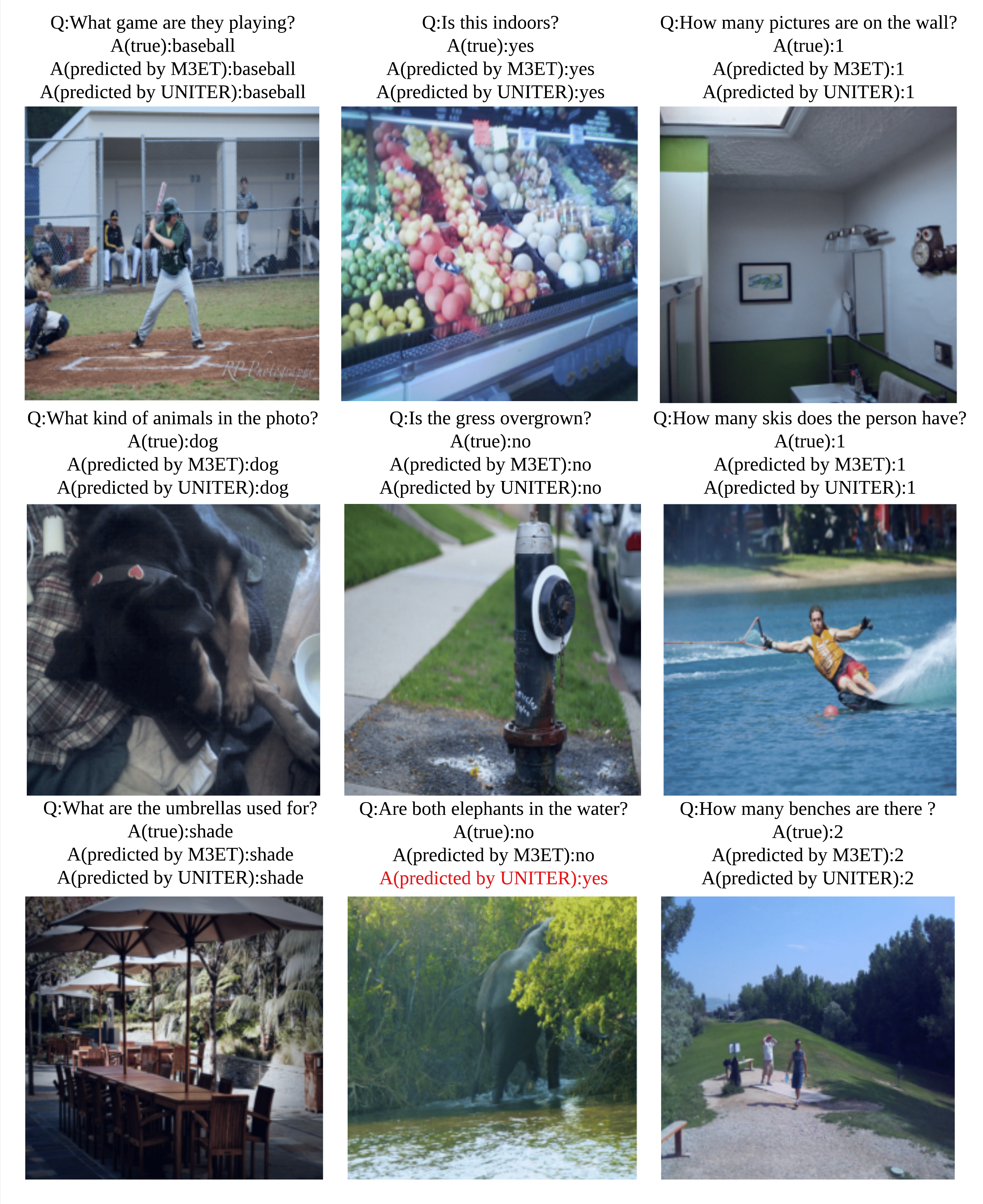}
    \caption{\small\textbf{VQA Task Visualization Example}}
    \label{fig:image2}
\end{figure}

Qualitative results further support the model's strong understanding of image-text relationships. As shown in Figure \ref{fig:image2}, M3ET successfully answers complex questions based on visual and textual cues, demonstrating its ability to make informed decisions from integrated multimodal input. These results confirm the robustness of M3ET's multimodal understanding and its capacity to tackle challenging VQA tasks with improved accuracy.

\subsection{Ablation Study}
To better understand the contributions of different components in M3ET, we conduct an ablation study examining the effects of removing or replacing key modules, including the text modality, Mamba Blocks, and cross-modal attention mechanism, and their impact on image reconstruction and VQA performance.

\subsubsection{Effect of Text Modality}

The text modality is crucial for enhancing image reconstruction and downstream multimodal tasks. As shown in Table \ref{tab:text_modality_removal}, removing the text modality reduces PSNR from 18.30 dB to 17.62 dB and VQA accuracy from 74.18\% to 70.1\%, highlighting its importance. Removing the text modality limits the model’s ability to leverage high-level contextual cues, which are essential for reasoning in complex VQA tasks.

\begin{table}[ht]
\centering
\caption{\small Effect of Text Modality on Performance and FLOPs}
\resizebox{\linewidth}{!}{
\begin{tabular}{lccc}
\toprule
\textbf{Setting} & \textbf{PSNR(dB)} & \textbf{VQA Acc(\%)} & \textbf{FLOPs($\times 10^9$)} \\

\midrule
Full M3ET & 18.30 & 74.18 & 9.08 \\
w/o Text Modality & 17.62 & 70.1 & 9.66 \\

\bottomrule
\end{tabular}
}
\label{tab:text_modality_removal}
\end{table}

\subsubsection{Impact of Mamba Blocks}
As shown in Table \ref{tab:mamba_blocks_removal}, Mamba Blocks provide a more efficient alternative in M3ET, improving performance by optimizing computational complexity compared to the standard Transformer layer of MAE. While PSNR increases to 18.30 dB and VQA accuracy improves to 74.18\%, Mamba Blocks reduce FLOPs by about 8.93\%, enhancing computational efficiency and optimizing resource usage for lightweight applications. In conclusion, Mamba Blocks improve M3ET's computational efficiency, demonstrating better adaptability in large-scale multimodal tasks with strict computational costs.

\begin{table}[ht]
\centering
\caption{\small Impact of Mamba Blocks on Performance and FLOPs}
\resizebox{\linewidth}{!}{
\begin{tabular}{lccc}
\toprule
\textbf{Setting} & \textbf{PSNR(dB)} & \textbf{VQA Acc(\%)} & \textbf{FLOPs($\times 10^9$)} \\
\midrule
MultiMAE & 16.69 & N/A & 10.32 \\
Full M3ET & 18.30 & 74.18 & 9.08 \\
w/o Mamba Blocks & 16.53 & 72.4 & 9.96 \\
\bottomrule
\end{tabular}
}
\label{tab:mamba_blocks_removal}
\end{table}

\subsubsection{Effect of Cross Attention}
Cross attention is crucial for aligning image and text representations. As shown in Table \ref{tab:cross_attention_removal}, disabling the dynamic cross attention mechanism causes a 5.3\% drop in VQA accuracy (from 74.18\% to 69.0\%), despite a slight reduction in PSNR. This emphasizes the role of cross attention in improving multimodal semantic alignment, especially for reasoning-intensive VQA tasks. Without this mechanism, the model struggles to correlate image regions with textual descriptions, resulting in worse question-answering performance.

\setlength{\tabcolsep}{4pt} 
\begin{table}[ht]
\centering
\caption{\small Effect of Cross Attention on Performance and FLOPs}
\resizebox{\linewidth}{!}{
\begin{tabular}{lccc}
\toprule
\textbf{Setting} & \textbf{PSNR(dB)} & \textbf{VQA Acc(\%)} & \textbf{FLOPs($\times 10^9$)} \\
\midrule
Full M3ET & 18.30 & 74.18 & 9.08 \\
w/o Cross Attention & 16.23 & 69.0 & 9.73 \\
\bottomrule
\end{tabular}
}
\label{tab:cross_attention_removal}
\end{table}

\subsection{Feasibility of Deployment on Robot}

\begin{figure}[htbp]
    \centering
    \includegraphics[width=\columnwidth]{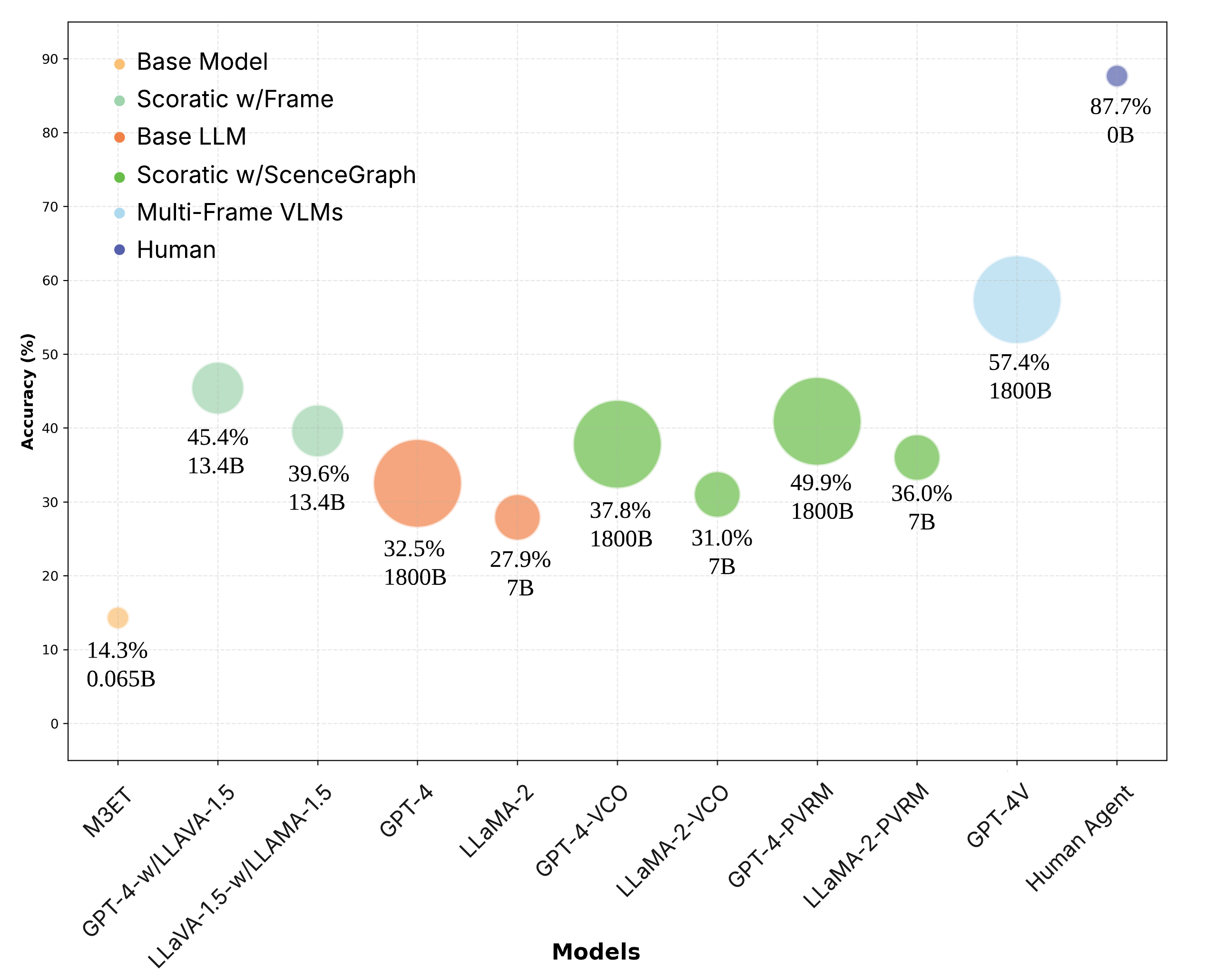}
    \caption{\small\textbf{Model Parameters and Accuracy comparison}}
    \label{fig:image5}
\end{figure}

We present M3ET, a multimodal architecture that integrates selective state space mechanisms to enhance model compression efficiency while retaining robust multimodal representation capabilities. As shown in Figure \ref{fig:image5}, M3ET employs a hybrid design that replaces parts of standard Transformer layers (7.09M parameters each) with efficient Mamba components (100.67K parameters each), reducing total parameters by 66.63\% from 195.97M (MultiMAE) to 65.39M. This framework achieves significant parameter reduction compared to large language models like GPT-4 (1.8T) and LLaMA-2 (70B). The optimized architecture results in 73.2\% lower memory consumption and 2.3 times faster inference, substantiating breakthrough efficiency gains.

\begin{figure}[htbp]
    \centering
    \includegraphics[width=\columnwidth]{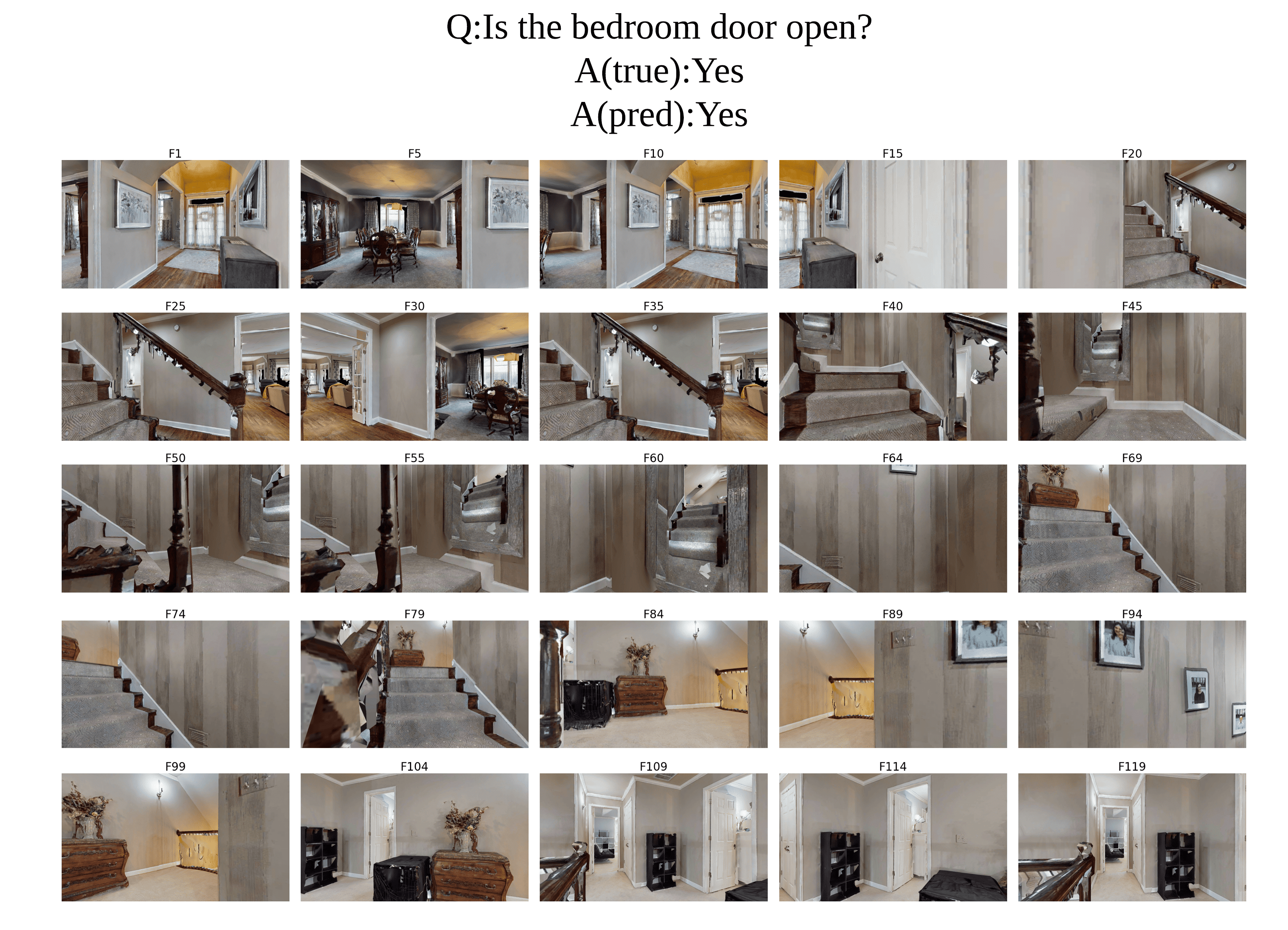}
    \caption{\small \textbf{EQA Task Visualization Sequence}}
    \label{fig:image4}
\end{figure}

In preliminary tests on the EQA task, as shown in Figure \ref{fig:image5}, the model's accuracy is 14.29\%, lower than state-of-the-art methods based on large language models. However, its lightweight nature is better suited for resource-constrained robot platforms. Figure \ref{fig:image4} shows multimodal reasoning examples of M3ET in typical EQA scenarios, validating its feasibility in robot vision-language tasks.

While our current evaluation uses standard datasets and simulated inputs, we plan to extend M3ET to real-world deployment on embedded robot platforms such as NVIDIA Jetson or ROS-based agents. This will allow us to validate inference latency, memory usage, and task performance in real-time conditions.

\section{Discussion}

The results obtained from our experiments show that M3ET is an efficient multimodal framework that significantly improves both efficiency and performance across various tasks. By reducing the model size by 66.63\% and increasing inference speed by 2.3×, M3ET demonstrates high computational efficiency while maintaining accuracy, particularly in Visual Question Answering (VQA) tasks, where it achieves an accuracy of 74.18\%. This efficiency is especially important for real-time applications on resource-constrained robotic platforms. However, despite these promising results, there are areas that can be further optimized. For example, M3ET's performance on the Embodied Question Answering (EQA) task remains limited, with an accuracy of only 14.29\%, which highlights the challenge of addressing complex open-ended reasoning tasks in dynamic environments. In the next phase of research, we will continue to explore ways to improve the model's performance in complex reasoning tasks and enhance its adaptability to diverse tasks.

In comparison with state-of-the-art lightweight multimodal methods, we believe that the large parameter sizes of existing models are too cumbersome for our application scenarios, particularly for resource-constrained robotic platforms. In contrast, M3ET's lightweight design makes it more suitable for these devices, enabling it to significantly reduce computational burden while retaining sufficient performance. Therefore, the lightweight and efficient nature of M3ET offers greater potential for deployment on real-world robotic platforms.

Although M3ET excels in multimodal fusion, its ability to handle more complex temporal dependencies, especially in long-sequence tasks, remains an area for future improvement. Future work will explore further techniques to reduce model size, enhance the model’s capability to handle complex reasoning tasks, and test M3ET in real-world environments to fully assess its deployment potential in robotics. The continued development of M3ET could lead to significant advancements in efficient multimodal learning, particularly in real-time robotic perception and decision-making applications.

\section{Conclusion}
This paper proposes M3ET, a lightweight multimodal framework that integrates semantic-guided attention with the Mamba module to enhance image-text fusion while reducing computational complexity. Validation across multiple datasets demonstrates that M3ET balances efficiency and accuracy, showcasing strong potential for real-time robotic applications. By improving the fusion of visual and textual information, M3ET enables more intuitive human-robot interactions, allowing robots to better understand and adapt to dynamic environments. Overall, M3ET enhances the robot's ability to perceive, interpret, and respond to human actions, improving performance in real-world settings. Future work will focus on optimizing M3ET's temporal modeling and multimodal fusion to further enhance performance in scenarios like EQA.

\section*{Acknowledgments}

The authors gratefully acknowledge the financial supports by different fundings. Kaixing Zhao is supported by the National Natural Science Foundation of China (No. 62407035), the China Postdoctoral Science Foundation (No. 2024M754226) and General Project of Taicang Basic Research Plan (No. TC2022JC11). Authors also thank the support of Hyper Creative Industry (Suzhou) Technology Co., Ltd.

\bibliographystyle{unsrt}
\bibliography{main}

\begin{thebibliography}{10}

\bibitem{ref2}
Li~Chen, Penghao Wu, Kashyap Chitta, Bernhard Jaeger, Andreas Geiger, and Hongyang Li.
\newblock End-to-end autonomous driving: Challenges and frontiers.
\newblock {\em IEEE Transactions on Pattern Analysis and Machine Intelligence}, 2024.

\bibitem{ref38}
Yunchao Tang, Shaojun Qi, Lixue Zhu, Xianrong Zhuo, Yunqi Zhang, and Fan Meng.
\newblock Obstacle avoidance motion in mobile robotics.
\newblock {\em Journal of System Simulation}, 36(1):1--26, 2024.

\bibitem{ref39}
Mohamed Reda, Ahmed Onsy, Amira~Y Haikal, and Ali Ghanbari.
\newblock Path planning algorithms in the autonomous driving system: A comprehensive review.
\newblock {\em Robotics and Autonomous Systems}, 174:104630, 2024.

\bibitem{ref40}
Tian Wang, Pai Zheng, Shufei Li, and Lihui Wang.
\newblock Multimodal human--robot interaction for human-centric smart manufacturing: a survey.
\newblock {\em Advanced Intelligent Systems}, 6(3):2300359, 2024.

\bibitem{ref41}
Ruo-Huai Sun, Xue Zhao, Cheng-Dong Wu, Lei Zhang, and Bin Zhao.
\newblock Research on mobile robot navigation method based on semantic information.
\newblock {\em Sensors}, 24(13):4341, 2024.

\bibitem{ref3}
Neelu Madan, Nicolae-C{\u{a}}t{\u{a}}lin Ristea, Kamal Nasrollahi, Thomas~B Moeslund, and Radu~Tudor Ionescu.
\newblock Cl-mae: Curriculum-learned masked autoencoders.
\newblock In {\em Proceedings of the IEEE/CVF Winter Conference on Applications of Computer Vision}, pages 2492--2502, 2024.

\bibitem{ref1}
Sara Atito, Muhammad Awais, and Josef Kittler.
\newblock Sit: Self-supervised vision transformer.
\newblock {\em arXiv preprint arXiv:2104.03602}, 2021.

\bibitem{ref42}
Peizhi Rong.
\newblock Ddnet: Depth dominant network for semantic segmentation of rgb-d images.
\newblock {\em Sensors}, 24(21):6914, 2024.

\bibitem{ref43}
Pan Zhang, Ming Chen, and Meng Gao.
\newblock Semantic guidance fusion network for cross-modal semantic segmentation.
\newblock {\em Sensors}, 24(8):2473, 2024.

\bibitem{ref6}
Yubo Zhang, Shuang Han, Zhongxin Zhang, Jianyang Wang, and Hongbo Bi.
\newblock Cf-gan: cross-domain feature fusion generative adversarial network for text-to-image synthesis.
\newblock {\em The Visual Computer}, 39:1283--1293, 2023.

\bibitem{ref12}
Jia Deng, Wei Dong, Richard Socher, Li-Jia Li, Kai Li, and Li~Fei-Fei.
\newblock Imagenet: A large-scale hierarchical image database.
\newblock In {\em 2009 IEEE conference on computer vision and pattern recognition}, pages 248--255. Ieee, 2009.

\bibitem{ref13}
Shengnan An, Zexiong Ma, Zeqi Lin, Nanning Zheng, and Jian-Guang Lou.
\newblock Make your llm fully utilize the context.
\newblock {\em arXiv preprint arXiv:2404.16811}, 2024.

\bibitem{ref44}
Tao Gong, Dan Chen, Guangping Wang, Weicai Zhang, Junqi Zhang, Zhongchuan Ouyang, Fan Zhang, Ruifeng Sun, Jiancheng~Charles Ji, and Wei Chen.
\newblock Multimodal fusion and human-robot interaction control of an intelligent robot.
\newblock {\em Frontiers in bioengineering and biotechnology}, 11:1310247, 2024.

\bibitem{ref15}
Paul~Pu Liang, Amir Zadeh, and Louis-Philippe Morency.
\newblock Foundations \& trends in multimodal machine learning: Principles, challenges, and open questions.
\newblock {\em ACM Computing Surveys}, 56(10):1--42, 2024.

\bibitem{ref45}
Qiongfeng Shi, Zhongda Sun, Xianhao Le, Jin Xie, and Chengkuo Lee.
\newblock Soft robotic perception system with ultrasonic auto-positioning and multimodal sensory intelligence.
\newblock {\em ACS nano}, 17(5):4985--4998, 2023.

\bibitem{ref46}
Wenhao Ding, Baiming Chen, Bo~Li, Kim~Ji Eun, and Ding Zhao.
\newblock Multimodal safety-critical scenarios generation for decision-making algorithms evaluation.
\newblock {\em IEEE Robotics and Automation Letters}, 6(2):1551--1558, 2021.

\bibitem{ref5}
Navid~Mohammadi Foumani, Chang~Wei Tan, Geoffrey~I Webb, and Mahsa Salehi.
\newblock Improving position encoding of transformers for multivariate time series classification.
\newblock {\em Data mining and knowledge discovery}, 38:22--48, 2024.

\bibitem{ref16}
Yitian Zhang, Liheng Ma, Soumyasundar Pal, Yingxue Zhang, and Mark Coates.
\newblock Multi-resolution time-series transformer for long-term forecasting.
\newblock pages 4222--4230, 2024.

\bibitem{ref17}
Fei Zhao, Chengcui Zhang, and Baocheng Geng.
\newblock Deep multimodal data fusion.
\newblock {\em ACM computing surveys}, 56(9):1--36, 2024.

\bibitem{ref18}
Hou-I Liu, Marco Galindo, Hongxia Xie, Lai-Kuan Wong, Hong-Han Shuai, Yung-Hui Li, and Wen-Huang Cheng.
\newblock Lightweight deep learning for resource-constrained environments: A survey.
\newblock {\em ACM Computing Surveys}, 56(10):1--42, 2024.

\bibitem{ref19}
Zhengqing Yuan, Zhaoxu Li, Weiran Huang, Yanfang Ye, and Lichao Sun.
\newblock Tinygpt-v: Efficient multimodal large language model via small backbones.
\newblock {\em arXiv preprint arXiv:2312.16862}, 2023.

\bibitem{ref20}
Yichen Zhu, Minjie Zhu, Ning Liu, Zhiyuan Xu, and Yaxin Peng.
\newblock Llava-phi: Efficient multi-modal assistant with small language model.
\newblock In {\em Proceedings of the 1st International Workshop on Efficient Multimedia Computing under Limited}, pages 18--22, 2024.

\bibitem{ref23}
Xing Wu, Guangyuan Ma, Meng Lin, Zijia Lin, Zhongyuan Wang, and Songlin Hu.
\newblock Contextual masked auto-encoder for dense passage retrieval.
\newblock 37:4738--4746, 2023.

\bibitem{ref27}
Bo~Peng, Eric Alcaide, Quentin Anthony, Alon Albalak, Samuel Arcadinho, Stella Biderman, Huanqi Cao, Xin Cheng, Michael Chung, Matteo Grella, et~al.
\newblock Rwkv: Reinventing rnns for the transformer era.
\newblock {\em arXiv preprint arXiv:2305.13048}, 2023.

\bibitem{ref29}
Yiheng Zhang, Ting Yao, Zhaofan Qiu, and Tao Mei.
\newblock Lightweight and progressively-scalable networks for semantic segmentation.
\newblock volume 131, pages 2153--2171. Springer, 2023.

\bibitem{ref10}
Tianwen Qian, Jingjing Chen, Linhai Zhuo, Yang Jiao, and Yu-Gang Jiang.
\newblock Nuscenes-qa: A multi-modal visual question answering benchmark for autonomous driving scenario.
\newblock In {\em Proceedings of the AAAI Conference on Artificial Intelligence}, volume~38, page 4542, 2024.

\bibitem{ref30}
Jihao Liu, Xin Huang, Jinliang Zheng, Yu~Liu, and Hongsheng Li.
\newblock Mixmae: Mixed and masked autoencoder for efficient pretraining of hierarchical vision transformers.
\newblock In {\em Proceedings of the IEEE/CVF Conference on Computer Vision and Pattern Recognition}, pages 6252--6261, 2023.

\bibitem{ref31}
Stanislaw Antol, Aishwarya Agrawal, Jiasen Lu, Margaret Mitchell, Dhruv Batra, C~Lawrence Zitnick, and Devi Parikh.
\newblock Vqa: Visual question answering.
\newblock In {\em Proceedings of the IEEE international conference on computer vision}, pages 2425--2433, 2015.

\bibitem{ref32}
Jiasen Lu, Dhruv Batra, Devi Parikh, and Stefan Lee.
\newblock Vilbert: Pretraining task-agnostic visiolinguistic representations for vision-and-language tasks.
\newblock {\em Advances in neural information processing systems}, 32, 2019.

\bibitem{ref33}
Hao Tan and Mohit Bansal.
\newblock Lxmert: Learning cross-modality encoder representations from transformers.
\newblock {\em arXiv preprint arXiv:1908.07490}, 2019.

\bibitem{ref34}
Yen-Chun Chen, Linjie Li, Licheng Yu, Ahmed El~Kholy, Faisal Ahmed, Zhe Gan, Yu~Cheng, and Jingjing Liu.
\newblock Uniter: Universal image-text representation learning.
\newblock In {\em European conference on computer vision}, pages 104--120. Springer, 2020.

\bibitem{ref35}
Hugo Touvron, Matthieu Cord, Matthijs Douze, Francisco Massa, Alexandre Sablayrolles, and Herv{\'e} J{\'e}gou.
\newblock Training data-efficient image transformers \& distillation through attention.
\newblock In {\em International conference on machine learning}, pages 10347--10357. PMLR, 2021.

\bibitem{ref36}
Mathilde Caron, Hugo Touvron, Ishan Misra, Herv{\'e} J{\'e}gou, Julien Mairal, Piotr Bojanowski, and Armand Joulin.
\newblock Emerging properties in self-supervised vision transformers.
\newblock In {\em Proceedings of the IEEE/CVF international conference on computer vision}, pages 9650--9660, 2021.

\end{thebibliography}


\end{document}